\documentclass[lettersize,journal]{IEEEtran}
\usepackage{amsmath,amsfonts}
\usepackage{algorithmic}
\usepackage[bookmarks=true]{hyperref}
\usepackage{graphicx}
\graphicspath{{Figures/}}
\usepackage{array}
\usepackage{textcomp}
\usepackage{stfloats}
\usepackage{url}
\usepackage{epstopdf} 
\usepackage{verbatim}
\usepackage{graphicx}
\def\BibTeX{{\rm B\kern-.05em{\sc i\kern-.025em b}\kern-.08em
		T\kern-.1667em\lower.7ex\hbox{E}\kern-.125emX}}
\usepackage{balance}
\usepackage{tabularx}
\usepackage{subfigure}
\usepackage{array}
\usepackage{color}
\usepackage{cite}
\usepackage{lipsum}
\usepackage{makecell}
\usepackage{enumerate}
\usepackage{tablefootnote}
\usepackage{amsthm}

\newtheorem{remark}{Remark}
\begin{document}
%

\title{A Compact Variable Stiffness Actuator \\for Agile Legged Locomotion}
\author{Lei Yu$^{1,2}$, 
	Haizhou Zhao$^1$, Siying Qin$^{1,2}$, Gumin Jin$^3$, and Yuqing Chen$^{1,*}$, \IEEEmembership{Member, IEEE}
	\thanks{
		This work was supported by the Jiangsu Science and Technology Program (BK20220283) and the Research Development Fund (RDF-20-01-08). \\
		\indent $ ^{1} $Authors are with Xi'an Jiaotong-Liverpool University, Suzhou, Jiangsu, China. 
		$ ^{2} $Authors are with University of Liverpool, United Kingdom. 
		$ ^{3} $Authors are with Department of Automation, Shanghai Jiao Tong University, Shanghai. \\
		\indent This paper has a supplement video material, available at http://ieeexplore.ieee.org.\\
		\indent $^*$E-mail: yuqing.chen@xjtlu.edu.cn
	}
}

\markboth{}%
{111}
\maketitle

\begin{abstract}

The legged robots with variable stiffness actuators (VSAs) can achieve energy-efficient and versatile locomotion. However, equipping legged robots with VSAs in real-world application is usually restricted by (i) the redundant mechanical structure design, (ii) limited stiffness variation range and speed, (iii) high energy consumption in stiffness modulation, and (iv) the lack of online stiffness control structure with high performance. In this paper, we present a novel Variable-Length Leaf-Spring Actuator (VLLSA) designed for legged robots that aims to address the aforementioned limitations. The design is based on leaf-spring mechanism and we improve the structural design to make the proposed VSA (i) compact and lightweight in mechanical structure, (ii) precise in theoretical modeling, and (iii) capable of modulating stiffness with wide range, fast speed, low energy consumption and high control performance. Hardware experiments including in-place and forward hopping validate  advantages of the proposed VLLSA.

\end{abstract}


\section{Introduction}

Agile legged locomotion, such as hopping, are characterized by the rapid release of a large amount of energy over a short duration during moving, making them a kind of explosive movements \cite{newton1996,wilson2003}. Variable stiffness actuators (VSAs) can change the output stiffness instantly with specific mechanical design, it is demonstrated that introducing VSAs could improve human–robot interaction, enable agile locomotion in legged robotic systems \cite{liu2019impedance,Sariyildiz2023}. Compactness of VSA design, including reduced installation size and light weight, plays a key role in improving the performance of agile locomotion tasks, as the compactness would improve the motion agility and energy efficiency \cite{Zhou2023,Chalvet2017_1}.



Integrating VSAs with elastic parts can be achieved through two configurations: series elastic actuation (SEA) and parallel elastic actuation (PEA) \cite{grimmer2012comparison}. SEA integrates the elastic part between the actuator and the load in series connection, offering several advantages such as improved error tolerance, reduced impact loads \cite{Yesilevskiy2015}, and precise torque measurement \cite{wu2020}. However, SEA demands large motor size or gear ratio in the rotary joints \cite{Wang2011}; this not only renders the robot's overall design redundant but also diminishes the prospects of achieving agile locomotion. 
Alternatively, PEA integrates an elastic element into an actuator by connecting it in parallel with the motor while directly coupling it to the load \cite{Ambrose2020}. By replacing a portion of the torque generated by the actuator with elastic components like springs, PEA can reduce the energy consumption of the actuators \cite{Batts2016} and its energy efficiency to SEA has been demonstrated before \cite{verstraten2016series,beckerle2016series,Sharbafi2019}. 
However, the effectiveness of PEAs is often constrained by the limited stiffness variation range and relatively high stiffness modulation power \cite{braun2019-2}. Besides, the inclusion of  clutches and additional mechanical components makes the structure bulky and heavy, while the stiffness modulation still couples with the robot configuration, both factors are especially noteworthy when dealing with tasks of agile locomotion\cite{Shin2023,Guenther2019}.

Recent studies have attempted to enhance energy efficiency and overcome the coupling by integrating PEA with switchable mechanism \cite{Chae2022agile}\cite{Xin2015}. However, such design still suffers from bulk structure and lacks online controlled stiffness output: the stiffness could only be changed statically during stance \cite{Hung2016}, restricting the dynamic control performance. Besides, PEA with leaf-spring, whose effective length is changed to modulate output stiffness, has theoretically guaranteed properties including infinite stiffness modulation range, low energy consumption and fast modulation speed \cite{Chalvet2018333}. PEAs with such design are commonly employed in prosthetic devices\cite{Shin2023,Shepherd2017,Sariyildiz2023} and wearable exoskeletons\cite{braun2019-2,David2018,Bergmann2022,Shao2024design, Hopkins2024quasi}. 
For instance, PEA with leaf-spring \cite{David2018} is controlled to imitate the swing phase of human locomotion, enabling knee retraction and extension; Combine leaf-spring mechanism with a nonlinear passive cam to create an energy-efficient biological hip joint motion during both the stance and swing phases\cite{shin2021power}. However, the proposed prototypes are generally bulky in size and weight, fail to implement agile legged locomotion; besides, the online stiffness modulation control structure is not proposed for highly dynamic tasks and not implemented in real-world experiments.

\begin{figure*}[ht!]
	\centering
	\includegraphics[width=\linewidth]{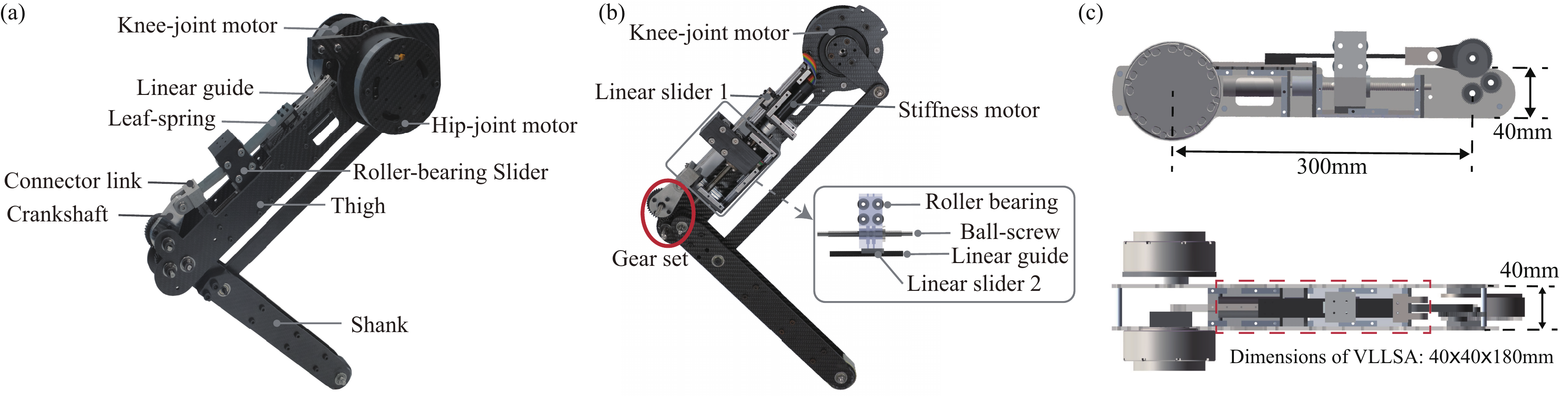}
	\caption{ (a) Overview of VLLSA-leg.   (b) Cross section of VLLSA-leg. (c) The installation size of VLLSA. The leg is controlled by the \textit{DJI Development Board Type C}  and the embedded brushless-FOC drivers on the joint motors. The controller board communicates with the motor drivers and broadcasts  commands through CAN at 1KHz. The hip and knee joints are respectively driven by 8016 motor with 1:6 planetary reducer. The stiffness motor is Maxon EC 22L(24V).
	}
	\label{fig:1}
\end{figure*}

In this paper, we propose a novel Variable-Length Leaf-Spring Actuator (VLLSA) for highly dynamic legged-robotic system, as shown in Fig. \ref{fig:1}. The proposed VLLSA belongs to the category of PEA, but overcomes the aforementioned limitations of existing VSA designs, enhancing the performance of legged robots in agile locomotion tasks. 
The VLLSA is designed with a direct connection to the knee joint, featured in introducing a slider with two pairs of roller bearings that could actively change the effective length of the elastic leaf-spring. 
This design enables convenient modification of the output stiffness, allowing for independent and versatile stiffness modulation to robot configuration. The endpoint of the leaf-spring is fixed to a linear guide, ensuring that the undeflected portion of the leaf-spring remains parallel to the movement of the slider.
To improve the control performance, the theoretical model of the VLLSA is analyzed and experimentally verified with experiments; besides, a real-time hopping control strategy considering variable stiffness modulation is proposed for agile locomotion tasks.  
The main contributions of this paper are:
%
%
%
%

\begin{enumerate}
	\item 	We present a novel design of VSA used in the legged robotic systems. Based on leaf-spring mechanism, the proposed VLLSA is designed to be compact in structure, the stiffness modulation is decoupled to the robot configuration, enabling convenient installation and versatile robot locomotion control.
	
	\item The theoretical model of the VLLSA is derived for open-loop control. The specific design of the VLLSA, i.e., employing linear guide and gear set,  ensures the precise modeling of the output stiffness with respect to the slider position and leaf-spring deflection angle, enabling accurate open-loop control of the output stiffness.

	\item We develop a real-time hopping control structure with online stiffness modulation for agile locomotion tasks. The control method is comprehensively tested in real-world experiments, where the VLLSA is demonstrated to provide output stiffness with wide range, fast modulation speed, and low energy consumption.
	
\end{enumerate}

The paper is structured as follows: In Section \ref{section2}, we introduce the mechanical design and mathematical model of the VLLSA. In Section \ref{verify_experiment}, we analyze the properties of VLLSA using the theoretical model and verify  with real-world experiments. In Section  \ref{section3}, we describe the hopping control strategy of the legged robot. In Section  \ref{section4}, we present the hardware experiment results. In Section  \ref{section5}, we discuss and conclude the paper.

\section{Variable Length leaf-spring Actuator}
\label{section2}
\subsection{Mechanical structure}


In this section, we present the prototype of a robotic leg equipped with VLLSA (VLLSA-leg), as is shown in Fig. \ref{fig:1}. The VLLSA-leg comprises hip and knee joints, both actuated by identical motors placed on the same axis to reduce the rotational inertia of the leg. The knee joint motor is used to help retract the leg to improve hopping performance. A rigid parallel four-bar structure is used to deliver the knee torque; the joint motors are direct torque-controlled. The joint angular feedback signals are obtained from the absolute encoders installed inside the motors. The main parts of the VLLSA-leg are 3D printed and the structure is reinforced with carbon fiber board to minimize the self-weight.

As shown in Fig. \ref{fig:1}a-b, one side of the leaf-spring in VLLSA is anchored to linear slider 1 while the other side is rigidly connected to the connector link hinged to the crankshaft, delivering the output stiffness to the knee joint via gear set. The ball-screw (pitch=2mm) driven by the stiffness motor is used to control position of the roller-bearing slider, which is supported by a linear guide to change the effective length of the leaf-spring, subsequently the output stiffness. 
We note that in similar VSAs using leaf spring \cite{liu2019impedance,braun2019-2},  
the effective length of leaf spring is also changed using ball screws, but movement of  the slider is supported by two cylinder linear guides, the roller blocks and guides are made with linear ball bearings; such design increases the radial dimensions of installation and self-weight. In contrast, our design constrains the roller-bearing slider to the ball screw and the linear slider 2, eliminating extra ball screw structure for compactness. 
Consequently, the proposed VLLSA is sufficiently compact to fit into the narrow inner space of the thigh leg (installation size shown in Fig. \ref{fig:1}c), while the external parts of VLLSA beyond the thigh segment do not affect motion of the leg. 
Furthermore, we note that changing stiffness requires low power when the device is at initial equilibrium state. Even without any electric energy input, the ball screw is capable of holding stiffness under a certain applied torque; meanwhile, the ball screw ensures that the motion of the roller block is back-drivable and energy-efficient.

\subsection{Theory guided design} 
\textbf{Gear set:} We aim to derive the mathematical model of VLLSA using Bernoulli-Euler beam theory \cite{shigley1972mechanical}. 
Large deflection in leaf spring usually leads to high nonlinearity, complicating the derivation of the output stiffness and consequently introducing modeling error; therefore, small deflection in leaf spring during robot operation is usually preferred. 
However, as the leaf spring is directly output to the knee joint, keeping small deflection in leaf spring would limit the operation range of knee joint and the dynamic performance of the legged robot. In our design, a custom made gear set is installed between the knee joint and the output of VLLSA, as marked in Fig. \ref{fig:1}b, the gear ratio is $ i:1:1 $. The gear set keeps the deformation of the active leaf spring $q$ within a small and safe range, while ensure the knee joint $\theta$ of the legged robot, marked in Fig. \ref{fig:2}c,  has enough range of motion, i.e., $\theta=\theta_0+q\cdot i $ where $\theta_0$ is the initial angle when leaf spring is undeflected.  The output torque of VLLSA $ \tau $ and the  gear-driven knee joint torque $\tau_{knee}$ has the relation as $ \tau=i\cdot \tau_{\text{knee}} $.

\textbf{Dual roller bearing \& Linear slider:} Another crucial factor that significantly affects the precision of stiffness estimation is the bending configuration of the leaf spring. For an ideal cantilever  support, it is essential that the undeflected segment maintains a zero slope at the contact point; in the VLLSA we employ the dual roller-bearing slider as the cantilever support, allowing movement along the length of the leaf spring \cite{braun2019-2,wu2020, Chaichaowarat2021}. While the two pairs of roller bearings restrict the vertical displacement of the spring at the  contact points, they do not ensure a zero slope along the undeflected length spring if the end of leaf spring is free, as depicted in Fig. \ref{fig:2}a. 
In contrast to the free-ended design \cite{braun2019-2}, in VLLSA, we attach the endpoint to a linear guide with a linear slider, as is illustrated in Fig. \ref{fig:2}b. Such design guarantees that the inactive portion of the leaf spring remains untwisted and consistently parallel to the motion of the roller-bearing slider. 

In summary, our distinctive design guarantees the small deflection of leaf spring, and consequently, the validity of the mathematical model derived from beam theory. The aforementioned design strategies are inspired by assumptions of the beam theory \cite{shigley1972mechanical}, and in turn, the theory guided design would improve the precision of the derived theoretical model of the VLLSA.

\subsection{Mathematical model} \label{model}
\begin{figure}
	\centering
	\includegraphics[width=0.9\columnwidth]{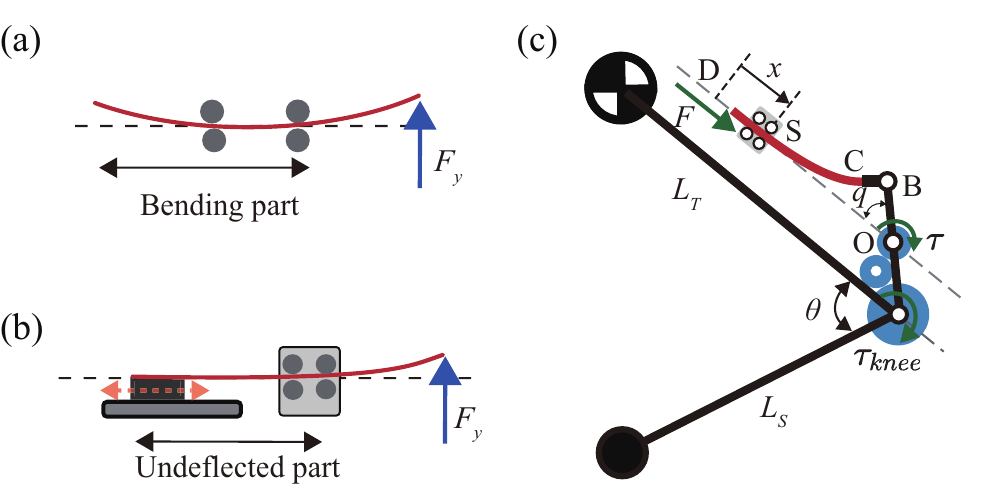}
	\caption{Modeling of VLLSA. (a) The bending part of leaf-spring within the two pairs of roller-bearing. (b) The slider and linear guide maintain a zero slope of the undeflected part. (c) The modeling mechanism of VLLSA-leg.  }
	\label{fig:2}
\end{figure}

The mechanism of the VLLSA-leg for modeling is depicted in Fig. \ref{fig:2}c: the red line represents the leaf-spring, the blue disks denote the gear set, the legged robot consists of the robot  thigh segment $ L_{T} $, shank segment $ L _{S} $, crank shaft BO, connect hinge BC, leaf spring CD, and a position-controlled slider S which adjusts the effective length of the leaf spring.  Assuming that the leaf-spring deforms in a small-deflection range, the output torque, output stiffness, and the force exerted by the stiffness motor to change the stiffness $ F  $  could be derived as \cite{braun2019-2}:
\begin{equation}
\tau(q, x) =-\frac{3 E I  e^{2}}{L^3} \frac{\cos q \sin q}{\left(\frac{a}{L}+1-\frac{x}{L}\right)^3-\left(\frac{a}{L}\right)^3},
\label{eq1} 
\end{equation}
\begin{equation}
K(q, x)  = \frac{3 E I  e^{2}}{L^3} \frac{\cos (2 q)}{\left(\frac{a}{L}+1-\frac{x}{L}\right)^3-\left(\frac{a}{L}\right)^3},
\label{eq2}
\end{equation}
\begin{equation}
F(q, x) =-\frac{9 E I e^2}{2 L^4} \frac{\left(\frac{a}{L}+1-\frac{x}{L}\right)^2 \sin ^2 q}{\left[\left(\frac{a}{L}+1-\frac{x}{L}\right)^3-\left(\frac{a}{L}\right)^3\right]^2} ,
\label{eq3}
\end{equation}
where $ L $ is the length of leaf-spring, $ E $ denotes  Young’s modulus, $ I $ denotes the area moment of inertia of the leaf spring, $ q $ is the leaf-spring deflection angle, $ \tau $ is the output torque, $ x $ is the slider position, $ e $ and $a$ are the length of OB and BC, i.e., the length of the lever arms connecting the spring to the output link.

The  models presented in \eqref{eq1}-\eqref{eq3} provide analytical expressions of output torque, stiffness and driven force with parameters of the deflection angle $ q  $ and the slider position $ x $. Effectiveness of the model with small  deflections in leaf spring has been verified in  \cite{braun2019-2}. In next section, we will analyze the theoretical model and verify the derived properties of the proposed VLLSA with real-world experiments.  
Specifications of the VLLSA-leg prototype are listed in Table \ref{tab:1}.

\begin{table}[ht!]
	\centering
	\caption{Specifications of VLLSA-leg}
	\label{tab:1}
	
	\setlength{\tabcolsep}{2.8mm}
	{
		\begin{tabular}{lccr}
			\hline\hline\noalign{\smallskip}	
			\textbf{Parameter} & \textbf{Symbol}&\textbf{Value}&\textbf{Unit}  \\
			\noalign{\smallskip}\hline\hline\noalign{\smallskip}
			Mass of  VLLSA-leg & $ M_{1} $&  $ 3.82 $& kg\\
			Mass of VLLSA\tablefootnote{The mass of VLLSA includes the leaf-spring made of 65Mn steel, ball screw, linear guide, roller-bearing slider, and stiffness motor} & $ M_{2} $&  $ 0.45 $& kg\\
			Mass of stiffness motor & $ m $&  $ 0.128 $& kg\\
			Hip/Knee joint motor & $ \tau_{1}/\tau_{2} $&  $ [0, 35] $& N$ \cdot $m\\
			Thigh/Shank segment length & $ L_{T}/L_{S} $& $ 0.35 $ & m\\
			Knee joint angle & $ \theta $& $ [50,  110] $& degree\\
			Leaf-spring deflection & $ q $& $ [0,  32] $&degree\\
			Gear ratio & $  i $& $ 1.87 $&-\\
			Initial knee angle & $  \theta_0 $& $ 50 $& degree\\
			Length of leaf spring & $L$ & $0.15$& m\\
						Width of leaf spring & $w$ & $0.018$ & m\\
			Young's modulus & $E$ & $196$ & GPa \\
			Area moment of inertia & $I$ & $2.4 \times 10^{-11}$ & m$ ^{4} $\\
			\noalign{\smallskip}\hline\hline
		\end{tabular}
	}  
\end{table}

\begin{remark} [Spring selection]\label{springselection}
    When deciding the spring specification, consider the output torque required for agile legged locomotion: when the leg crouches to store elastic energy, the output torque from VLLSA should be smaller than the nominal torque of joint motor, i.e., 8Nm, to avoid reverse output from joint motor; when the leg retracts, the output torque should be sufficiently small to save retraction motor torque. 	
	Considering the model derived in \eqref{eq1}, we note that parameters $E$, $I$, and $L$ are decided by leaf-spring, where $I=nwh^3/12$, $n$, $w$, and $h$ represent the number of stacked spring pieces, width, and thickness of the leaf-spring. While length $L$ and width $w$ of the spring are limited by the mechanical design, thickness $h$ is determined by the standard type from manufacturer, we select the metal material of leaf-spring as 65MN steel \cite{braun2019-2} and the Young's modulus $E$ is thus decided. Besides, for small deflection angle $q\in[0,20^\circ]$, it is computed from \eqref{eq1} that $n=2$ satisfies the demands of agile locomotion tasks.

\end{remark}

\section{Experimental Validation of VLLSA model} \label{verify_experiment}
\subsection{Test platform}
\begin{figure}[ht!]
	\centering
	\includegraphics[width=0.8\linewidth]{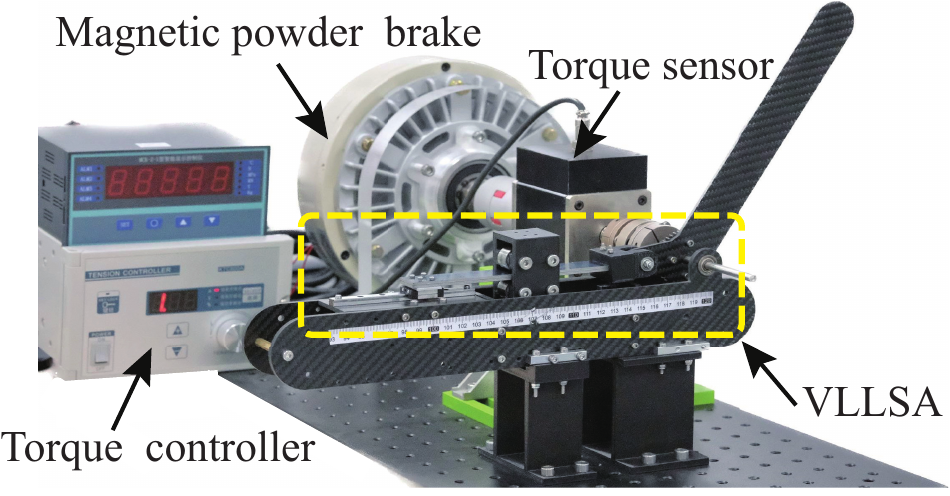}
	\caption{The experimental platform to test VLLSA. }
	\label{fig:3}
\end{figure}

To validate properties of the proposed VLLSA, we have built an experimental platform, as is shown in Fig. \ref{fig:3}. The output shaft of the VLLSA is connected to a torque sensor with the measuring range of [0, 50] Nm; a magnetic powder brake is connected to the other side of the torque sensor to provide resistance moment. This test platform will  be used to test the stiffness modulation range in Section \ref{range}, stiffness modulation speed in Section \ref{speed} and power consumption of VLLSA in Section \ref{power}. 

\subsection{Output torque and stiffness} \label{range}
\textbf{Theoretical Analysis:} Based on the  nonlinear relations of the  torque and stiffness to the slider position in \eqref{eq1}-\eqref{eq2}, the proposed design can output a wide range of torque and stiffness by changing the effective length of the leaf-spring. Taking stiffness for instance, when the slider is close to hip, 
\begin{equation}
K_{\min } =  K(q, 0) = \frac{3 E I  e^{2}}{L^3} \frac{\cos (2 q)}{\left(\frac{a}{L}+1\right)^3-\left(\frac{a}{L}\right)^3}, 
\label{eq4}
\end{equation}
given $0<e \ll L$, we have $K_{\min } \approx 0$,  i.e., the output stiffness is close to zero. When the slider is close to the knee joint, i.e.,  $ x \rightarrow L $, 
\begin{equation}
\label{kmax}
K_{\max } = \lim _{x \rightarrow L} K(q, x )\rightarrow \infty,  
\end{equation}
the output stiffness tends to be infinite, and the output link becomes rigidly connected to the frame of the actuator. Therefore, given $0<e \ll L$, the  output stiffness range is:
\begin{equation}
\label{eq6}
K(q, x) \in\left[K_{\min }, K_{\max }\right)\approx\left[0, \infty\right). 
\end{equation}
Similar conclusion also applies to the output torque in \eqref{eq1}.

 \begin{figure*}[t]
 	\centering
 	\includegraphics[width=\textwidth]{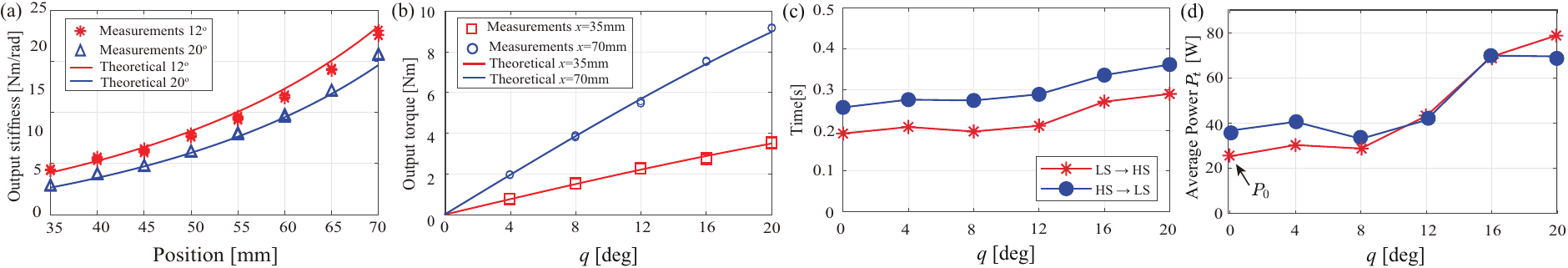}
 	\caption{Experiment results. (a) Output stiffness with respect to slider position. Two fixed deformation angles of leaf spring $q= \{12^{\circ}, 20^{\circ}\}$ are selected for demonstration. The red star and blue triangles denote the data with $q=12^{\circ}$ and $q=20^{\circ}$ respectively.
 (b) Output torque with respect to deformation angle. The red square and blue circle denote the data with $x=35$mm and $x=70$mm respectively.
 		(c) Speed of output stiffness modulation. The red star and blue dot denote the data with variation of Low Stiffness (LS, $ x=35$ mm) to High Stiffness (HS, $ x=70$ mm) and HS to LS respectively.
 		(d) The average power to change the output stiffness, the legend  is the same as in (c).}
 	\label{fig:4}
 \end{figure*}

\textbf{Experimental Validation:} 
To test the stiffness modulation range in VLLSA, we select two  deflection angles $ q= 12^{\circ}, 20^{\circ} $ within the operational range of leaf spring for instance. 
Within a certain range of slider movement, if $ x $ is too small, the stiffness variation will  be insignificant; if $x$ is too large, the stiffness variation will be difficult to measure. Taking into account the practical structural design, under constant $q$, the roller slider is moved  from $ x = 35 $ mm to $ 70 $ mm for measurement, during which the output stiffness and slider position are recorded. The experiment is repeatedly conducted  three times to ensure the consistency.

As  shown in Fig. \ref{fig:1}a, it is clear that the measured output stiffness  aligns with the theoretical predictions from \eqref{eq2}. Within the high-stiffness domain, significant stiffness variations can be achieved by changing the effective spring length. For instance, when the deflection angle of the leaf-spring is $ 12^{\circ} $ and the slider moves from $ x = 35  $ mm to $ x = 70$ mm, the output stiffness increases from 9.43 Nm/rad  to 22.55 Nm/rad, by a ratio of 239\%. 
We also validate the relationship between the output torque $\tau$ and deflection angle $q$. We select $ q  =0^\circ$, $4^\circ$, $8^\circ$, $12^\circ$, $16^\circ$, and $20^\circ$, with the effective lengths of the leaf-spring set at boundary value $x = 35, 70$ mm. As shown in Fig. \ref{fig:4}b, the output torque $\tau$ increases after $q$, and aligns with the theoretical prediction from \eqref{eq1}. 
Besides, we note that the three repeated trials shows high consistency, that the measured data are almost overlapped at every measurement point; such consistency ensures the robustness of the open-loop control.

While the alignment between experimental data and theoretical predictions is observed, we also note that there are small error between the measured and model-predicted results, especially when the slider approaches to the knee joint, i.e., $x\in[60,70]$ mm. This might be caused by the inevitable nonlinearity in the leaf spring, manufacturing error and the deformations in delicate system components.  However, despite the small error, the legged robot can still operate successful hopping locomotion with open-loop controlled VLLSA, demonstrating the validity and robustness of the proposed model in \eqref{eq1}-\eqref{eq2}.

\subsection{Stiffness modulation speed}  \label{speed}
\textbf{Theoretical Analysis:} To simplify the analysis, we assume that the deflection angle $q$ is constant, i.e., $\dot{q}=0$, the speed of stiffness modulation $v$ could be expanded as
\begin{equation}
v(q,x) = \frac{d K(q, x)}{d t}\bigg|_{\dot{q}=0}=\frac{d K(q, x)}{d x} \cdot \dot{x},  
\label{eq5}
\end{equation}
where the first term, $dK/dx$, is determined by the mechanical design and could be derived from equation (\ref{eq2}), and the second term, $\dot{x}$, simply refers to the movement speed of the slider driven by the ball screw, then we can rewrite \eqref{eq5} as
\begin{align}
v(q,x) =\frac{9 E I e^2}{L^4} \frac{ \left(\frac{a}{L}+1-\frac{x}{L}\right)^2\cos (2 q)}{[\left(\frac{a}{L}+1-\frac{x}{L}\right)^3-\left(\frac{a}{L}\right)^3]^{2}}\;\cdot\;\dfrac{np}{60},
\label{eq9}
\end{align}
where $ p $ is the lead of ball-screw, denoting the displacement of the slider when the motor rotates one circle, and $ n $ is the rotational speed of stiffness motor.

Analyzing equation \eqref{eq9}, we find that when the slider moves close to knee joint, i.e., $x\rightarrow L$, equation \eqref{eq9} gives
\begin{equation}
v(q,x) =  \lim _{x \rightarrow L} \frac{d K(q, x)}{d t} \rightarrow \infty,
\end{equation}
i.e., the speed of stiffness modulation keeps increasing to infinitely when the slider approaches to knee joint. 

Besides, we note that when the mechanical design and the configuration of legged robot are fixed, if $q$ is sufficiently small,  $v$ in \eqref{eq9} is linearly proportional to $n$, the rotational speed of stiffness motor. This provides an intuitive method to control the stiffness modulation speed: controlling the speed of stiffness motor could control the stiffness modulation speed, which is significant to legged robot in agile locomotion  \cite{chen2022}.

\textbf{Experimental Validation:} 
We mainly test the consistency in repetitive stiffness modulation, which is vital for the legged robotic systems. The stiffness motor is controlled at 2000 rpm with a constant voltage of 24 V. We select several sampled deflection angles $q = 0^\circ, 4^\circ, 8^\circ, 12^\circ, 16^\circ, 20^\circ $, to maintain the constant deflection during measurement, the spring is loaded by the magnetic powder brake via the torque sensor, as shown in Fig. \ref{fig:3}; under each angle, we control the slider to move back and forth between Low Stiffness (LS, $ x=35$ mm) and High Stiffness (HS, $ x=70$ mm) modes, and measure the stiffness modulation time, starting from the commanded start of motor and ending when the slider reaches the target position. 
The experiment results are presented in Fig. \ref{fig:4}c. 


We first observe that the stiffness variation of HS$\rightarrow$LS takes  longer time than LS$\rightarrow$HS, this is mainly due to the measurement mechanism: the starting of measurement is set at the stiffness motor is commanded to rotate, and because the static friction of HS is larger than LS due to the pressure from leaf spring,  the stiffness motor would take longer time to move the slider. Besides, it is clearly shown in Fig. \ref{fig:4}c that the stiffness modulation speed is consistent across the same operation: when the deflection angle is small, $q\leq12^\circ$, the time spent in HS$\rightarrow$LS and LS$\rightarrow$HS is around 0.28s and 0.2s; when the deflection angle is larger, $12^\circ\leq q\leq20^\circ$, the stiffness modulation time increases to around 0.35s and 0.29s due to the increased resistance from leaf spring. 


\begin{table*}[h]
	\centering
	\caption{ Performance of existing actuators}
	\label{tab:2}
	\setlength{\tabcolsep}{2.5mm}{\begin{tabular}{p{2.3cm}p{1.8cm}p{2.5cm}p{2.2cm}p{2.2cm}p{0.6cm}p{2cm}}
			\hline\hline\noalign{\smallskip}	
			\textbf{Actuator} & \textbf{Stiffness range [Nm/rad]}&\textbf{Average speed of stiffness modulation [Nm/(rad$\cdot$s)]}&\textbf{Stiffness modulation power $P_t$ [W]}&\textbf{Static power to hold stiffness [W]}&\textbf{Weight [Kg]}&\textbf{Installation size \hspace{1cm}[$ \textup{mm}^{3} $]}  \\
			\noalign{\smallskip}\hline\noalign{\smallskip}
			VLLSA & [0, 1787]&  2961.7 & 24.5 & 1.20 & {0.45}& {$ 180\times40\times40 $}\\
			HVSA \cite{kim2012design} & [0, 120]&  750 & - & - & 1.80& $ 50^{2}\times \pi\times258 $\\
			vsaUT-II\cite{Groothuis2014} & [0.7, 948] &  1215 &  69.8& - & 2.50& -\\
			AwAS\cite{Jafari2013} & [30,1500]&  420 &  [2.5,15] & - & 1.80 & $ 270\times130\times130 $\\
			PVSA\cite{mathews2022design} & [1.9, 29.3]&  28.7 & 52 & -& 1.90& $ 152^{2}\times \pi\times 142 $\\
			VSSA\cite{braun2019-2}& {[10, 7800]}&  9990& {22.0}&{2.25}& 3.00&$ 360\times100\times90 $ \\
			Swi-CA\cite{Shin2023} & [0.15, 3873]&  6504& - & {1.64}& 2.36& $ 210\times163\times117 $ \\

			\noalign{\smallskip}\hline\hline
	\end{tabular}}  
	
\end{table*}

\subsection{Stiffness modulation power} \label{power}
\textbf{Theoretical Analysis:} 
The total electrical power $ P_{t} $ required to modulate the stiffness can be expressed as 
\begin{equation}
P_{t} = P_{0} + \Delta P,
\label{eq10_1}
\end{equation}
where $P_0$ is the power to drive the slider when no load, and  $\Delta P$ is the power to overcome the resistance from the leaf-spring's deformation and hold the stiffness. We note that $P_t$ includes the power to move the slider, deform the leaf spring and hold the stiffness.

According to the ball-screw mechanism \cite{wei2011kinematical}:
\begin{equation}
\begin{aligned}
P_{0} = F_{r} \dot{x}
      =  \dfrac{2 \pi \eta \tau_s}{p} \cdot \dfrac{np}{60}
      = \dfrac{\pi \eta n \tau_s}{30} 
\end{aligned},
\end{equation}
where $ F_{r} $ is the output force of stiffness motor, $ \eta $ is the efficiency of the ball-screw driven system, $ \tau_s $ is the output torque of stiffness motor; $ P_{0} $ is determined by $ \tau_{s} $ and $ n $. In our particular configuration, the motor stiffness operates at a constant rotational speed, and the power demand of the stiffness motor depends on its rotational speed.

Based on \eqref{eq3}, the  power required to hold the stiffness is
\begin{equation}
\Delta P(q, x) = F(q, x) \dot{x}, 
\label{eq10}
\end{equation}
As the function (\ref{eq3}) indicates, when the robot leg extends to the initial equilibrium configuration, i.e., $ q = 0^\circ $, $F(0, x)=0$, then the required power to hold the stiffness is
\begin{equation}
\Delta P_{\text{min}} = F(0, x) \dot{x} =0.  
\end{equation}
This feature ensures that, at the initial equilibrium configuration, the motor can adjust the stiffness of the actuator without being opposed by the spring. 
When the robot leg is not at the initial equilibrium configuration, i.e., $ q \neq  0^\circ $, the actuator's ability to achieve infinite-range stiffness modulation is primarily constrained by its physical limitations.
It is clear that the stiffness is proportional to the slider position. When $ 0 \leq x \leq x_{\text{max}} \in [0,L] $,  the maximum  power required to hold the stiffness is 
\begin{equation}
\Delta P_{\text{max}}=F(q, x_{\text{max}} ) \dot{x}. 
\end{equation}

Therefore, the total electrical power $ P_{t} $ in \eqref{eq10_1} is bounded as
\begin{equation}
P_{t} \in [\dfrac{\pi \eta n \tau_s}{30}, \dfrac{\pi \eta n \tau_s}{30} + \dfrac{np}{60} F(q, x_{\text{max}})]\subset(0,\infty),
\end{equation}
it indicates that $P_{t}$ remains bounded regardless of the spring deflection $q$ and slider position $x$. 
It is clear that the VLLSA modulate a wide range of output stiffness while require only bounded modulation electric power.

%
%

\textbf{Experimental Validation:}  The experiment we conducted to verify the stiffness modulation power is to measure the consumed electric power of stiffness motor, as it is the only part that requires power input to VLLSA. The test is to measure the consumed power when the slider is moving back and forth between Low Stiffness (LS, $ x=35$ mm) and High Stiffness (HS, $ x=70$ mm), as is the setting when measuring the stiffness modulation speed in Section \ref{speed}. 


When  the leaf-spring is at its initial equilibrium position $q = 0^{\circ}$, the Maxon motor drives the slider from LS to HS (marked by the red star in Fig. \ref{fig:4}d), the baseline electrical power $ P_{0} $ is calculated to be  $ 23.48 $ W (marked in Fig. \ref{fig:4}d). 
Compare with LS $ \rightarrow $ HS, the slider moves from HS $ \rightarrow $ LS (marked by the blue circle in Fig. \ref{fig:4}d), the power consumed is higher because the slider should overcome the static friction.
We notice that the angle $ q $ varies within the range of $ [0^\circ, 12^\circ] $, the power required for stiffness modulation is limited and remains nearly constant,  the result is consistent with the $ F $ in \ref{eq3}. When $ q \in [12^\circ, 20^\circ]  $, the angle is relatively large, regardless of the slider moves forward or backward, the total power $ P_{t} $ increases by a factor of $ \approx 75 \% $, accounting for the energy needed to maintain stiffness and induce spring deformation.


\subsection{Comparison}
To highlight the advantages of VLLSA, we compare with other state-of-the-art alternatives, in terms of stiffness modulation range, speed, power consumption, weight and size. The results are summarized in  Table \ref{tab:2}. 

We first notice that when compared to other alternatives \cite{kim2012design,Groothuis2014,Jafari2013} and recent legged PEA  \cite{mathews2022design}, the proposed VLLSA, VSSA \cite{braun2019-2} and the very recent Swi-CA \cite{Shin2023} could modulate the output stiffness with wider range, higher speed and lower power consumption, due to the employment of leaf-spring mechanism. The VSSA \cite{braun2019-2} provides larger output stiffness by stacking around 10 pieces of leaf-springs together; while the  VLLSA could also realize similar range with more springs, we select only 2 pieces to meet the requirements for agile locomotion, as is analyzed in Remark \ref{springselection}. The Swi-CA \cite{Shin2023} demonstrates rapid  modulation speed in highly nonlinear range of stiffness output,  such feature is difficult to be utilized in practical applications; but the  VLLSA could achieve fast modulation speed with relatively linear relation, convenient for robot controllers. Furthermore, we note that VLLSA has the most compact size and lightest weight among all other alternatives, such advantages would equip the legged robot with reduced inertia to improve the  performance in agile and highly dynamic locomotion tasks \cite{Zhou2023}.

\textbf{Summary:} In this section, it is demonstrated that the models \eqref{eq1}-\eqref{eq2} given in Section \ref{model} could provide precise modeling of the output stiffness and torque of the proposed VLLSA, as is demonstrated in Section \ref{range}; in Section \ref{speed}-\ref{power}, the analytical properties of the stiffness modulation speed and power are derived, analyzed and verified with hardware experiments. We have also compared the proposed VLLSA with other alternatives to highlight our advantages: \textit{with compact installation size and self-weight, the VLLSA could provide sufficiently large stiffness modulation range, fast modulation speed, while consume low power}, making it suitable for agile locomotion tasks.

Besides, it is revealed in Fig. \ref{fig:4}c-d that the VLLSA has consistent and energy-efficient performance when $ q $ is under small deflection, i.e., $q\in[0,12]^\circ$. As $ q $ changes within $ [0,12]^\circ $, correspondingly, the knee joint angle $ \theta  \in[50,72.44]^\circ$  due to the amplification of gear set. In the real-world experiment, we aim to control the motion such that the robot operates almost within this range to achieve the best dynamic performance.

\begin{figure*}[]
	\centering
	\includegraphics[width=\textwidth]{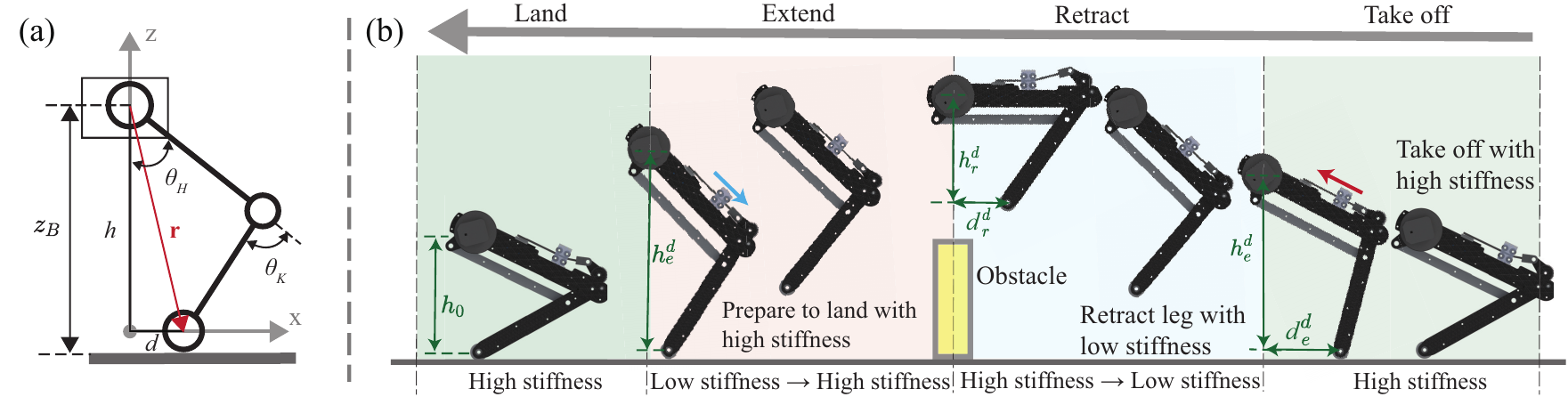}
	\caption { Hopping control strategy. (a) Parameters of the VMC designed for VLLSA-leg; (b) Hopping control strategy. The red arrow shows slider switches from high stiffness to low stiffness, and the blue arrow shows slider switches from low stiffness to high stiffness.  }
	\label{fig:5}
\end{figure*}

\section{Real-time Hopping Controller}
\label{section3}
\subsection{Virtual model control}
We have presented the design and dynamic properties of the VLLSA in Sections \ref{section2}--\ref{verify_experiment}. 
To test the performance of VLLSA in legged robot, we implement the typical virtual model control (VMC) \cite{pratt2001virtual} to formulate a PD-like feedback controller for hopping task. In this section, we will briefly present the VMC method formulated for the legged robot system with VLLSA. 
We define the following generalized coordinate variable $\mathbf{q}\in\mathbb{R}^3$ for the legged robot system with VLLSA: $\mathbf{q} = [z_{B} \quad \theta_{H} \quad \theta_{K}]^\top$, 
where $z_{B}$ is the base height from the ground, $\theta_H$ is the hip joint angle and $\theta_K$ is the knee joint angle, as are shown in Fig. \ref{fig:5}a. 

In the repetitive hopping task, the legged robot dynamics could be divided into two phases: stance and flight. We first consider the dynamics in stance mode, i.e., the foot is in contact with the ground. The contact dynamics of the system can be expressed in the closed form as
\begin{equation}
\begin{aligned}
\mathbf{M}(\mathbf{q})\ddot{\mathbf{q}}+\mathbf{H}(\mathbf{q}, \dot{\mathbf{q}})=\begin{bmatrix}0 \\ \boldsymbol{\tau}_J\end{bmatrix} + \mathbf{J}_c^T\mathbf{F}_c , \\
\text{subject to}\quad \mathbf{J}_c(\mathbf{q})\ddot{\mathbf{q}} + \dot{\mathbf{J}}_c(\mathbf{q}, \dot{\mathbf{q}})\dot{\mathbf{q}} = \mathbf{0},
\end{aligned}
\label{dynamics}
\end{equation}
where $\mathbf{M}\in \mathbb{R}^{3\times3}$ is the joint space inertia matrix, $\mathbf{H}\in \mathbb{R}^{3}$ is the collection of nonlinear effects including Coriolis and generalized gravity, $\boldsymbol{\tau}_J\in\mathbb{R}^2$ is the output torque from hip and knee joints, $\mathbf{J}_c\in \mathbb{R}^{3\times3}$ is the contact Jacobian matrix and $\mathbf{F}_c\in \mathbb{R}^3$ is the point-contact force.

To control the robot, we implement the VMC method which regards the robot as a spring-damper system with respect to  the state variables
\begin{equation}\label{vmcstate}
\mathbf{r}=[d, h]^T, \;\;\mathbf{\dot{r}} = [\dot{d}, \dot{h}]^T ,   
\end{equation}
where $d, h$ are respectively x and z-axis distance of the foot contact frame with respect to the hip frame, as shown in Fig. \ref{fig:5}a. Input to the spring-damper system is a virtual force 
\begin{equation}
\label{vmcforce}
\mathbf{F}_v=\mathbf{K}_p(\mathbf{r}^d - \mathbf{r}) + \mathbf{K}_d(\dot{\mathbf{r}}^d-\dot{\mathbf{r}}) ,
\end{equation}
where $\mathbf{F}_v\in\mathbb{R}^{2}$ is the virtual force exerted in x and z-axis direction, $\mathbf{r}^d$ is the desired position, and $\mathbf{K}_p, \mathbf{K}_d\in\mathbb{R}^{2\times2}$ are two  matrices defining the stiffness and damping of the virtual model. 
The VMC controller \cite{pratt2001virtual} sets the joint torques to 
\begin{equation}\label{eq:vmc_tq}
\boldsymbol{\tau} _J= [\mathbf{0}_{2\times1}\quad\mathbf{I}_{2\times2}](\mathbf{H}(\mathbf{q}, \dot{\mathbf{q}}) + \mathbf{J}_v^T\mathbf{F}_v - \mathbf{J}_c^T\hat{\mathbf{F}}_c) , 
\end{equation}
where $\mathbf{J}_v\in\mathbb{R}^{2\times3}$ is the virtual contact point Jacobian and $\hat{\mathbf{F}}_c$ is the estimated contact force. For convenience, we set the virtual contact point to be the same as the foot contact point, i.e., $\mathbf{J}_v=\mathbf{J}_c$, and the $\hat{\mathbf{F}}_c$ is set as $[0, 0, M_1g]^T$ where $g$ is the gravity of earth. Substituting (\ref{eq:vmc_tq}) into (\ref{dynamics}), one can derive the contact dynamics with VMC controller:
\begin{equation}
\mathbf{M}(\mathbf{q}) \ddot{\mathbf{q}} + \begin{bmatrix}
H_b(\mathbf{q}, \dot{\mathbf{q}}) \\ \mathbf{0}_{2\times1}\end{bmatrix}=\begin{bmatrix}
\mathbf{J}_{cb}^T \mathbf{F}_{c}\\
\mathbf{J}_{cj}^T(\mathbf{F}_{v} + \mathbf{F}_c - \hat{\mathbf{F}}_c)
\end{bmatrix}, 
\end{equation}
where 
$H_b\in\mathbb{R}$ is the base nonlinear effect, i.e., Coriolis force and gravity, $\mathbf{J}_{cb}\in\mathbb{R}^{2\times1}$ and $\mathbf{J}_{cj}\in\mathbb{R}^{2\times2}$ are the base and joint columns of the contact jacobian. Assuming accurate approximation, i.e., $\hat{\mathbf{F}}_c\approx\mathbf{F}_c$, the virtual model dynamics as a function of $(\theta_H, \theta_K)$ is then controlled by the PD-like virtual contact force in (\ref{vmcforce}), simulating a spring-damper system where the reactive force is linearly proportional to the displacement and velocity.


When the robot is off-ground in flight mode, the contact force is zero and the VMC-controlled dynamics could be simplified as
\begin{equation}\label{vmcload}
\mathbf{M}(\mathbf{q}) \ddot{\mathbf{q}} + \begin{bmatrix}
H_b(\mathbf{q}, \mathbf{\dot{q}}) \\ \mathbf{0}_{2\times1}\end{bmatrix}=\begin{bmatrix}
0\\
\mathbf{J}_{cj}^T \mathbf{F}_{vj}
\end{bmatrix}. 
\end{equation}

We note that the torque applied to knee joint combines the output from both VLLSA and motor, thus saving the power of the knee joint motor. Consequently, to implement the VMC method, the torque required from the hip and knee motors $\mathbf{\tau}_M\in\mathbb{R}^2$ could be computed by
\begin{equation}
	\label{controller}
\boldsymbol{\tau}_M =  \boldsymbol{\tau} _J-[0,\tau_{knee}]^\top
\end{equation}
where $\tau_{knee}$ is the output torque of VLLSA at knee joint.

One of the advantages when using the VMC controller in \eqref{eq:vmc_tq} is that it only requires the feedback parameters $d$ and $h$, both of which could be evaluated directly using the angles $\theta_H$ and $\theta_K$ measured by the encoder in the joint motor. Consequently, the height $z_B$ is not required by the controller, eliminating the influence from environment and making the VLLSA-leg a self-contained robotic system.


We note that the principles of VMC assume the model to have concentrated inertia on the non-rotating link; however, the slider in VLLSA cannot guarantee such assumption. As will be shown later in the experiments, the robot cannot precisely control the x-axis position, i.e., $d$ shown in Fig. \ref{fig:5}, during the highly dynamic aerial phase of the hopping; but as the paper is to demonstrate the effectiveness of VLLSA, such deviation is relatively acceptable.
 \begin{figure*}[]
	\centering
	\includegraphics[width=0.9\linewidth]{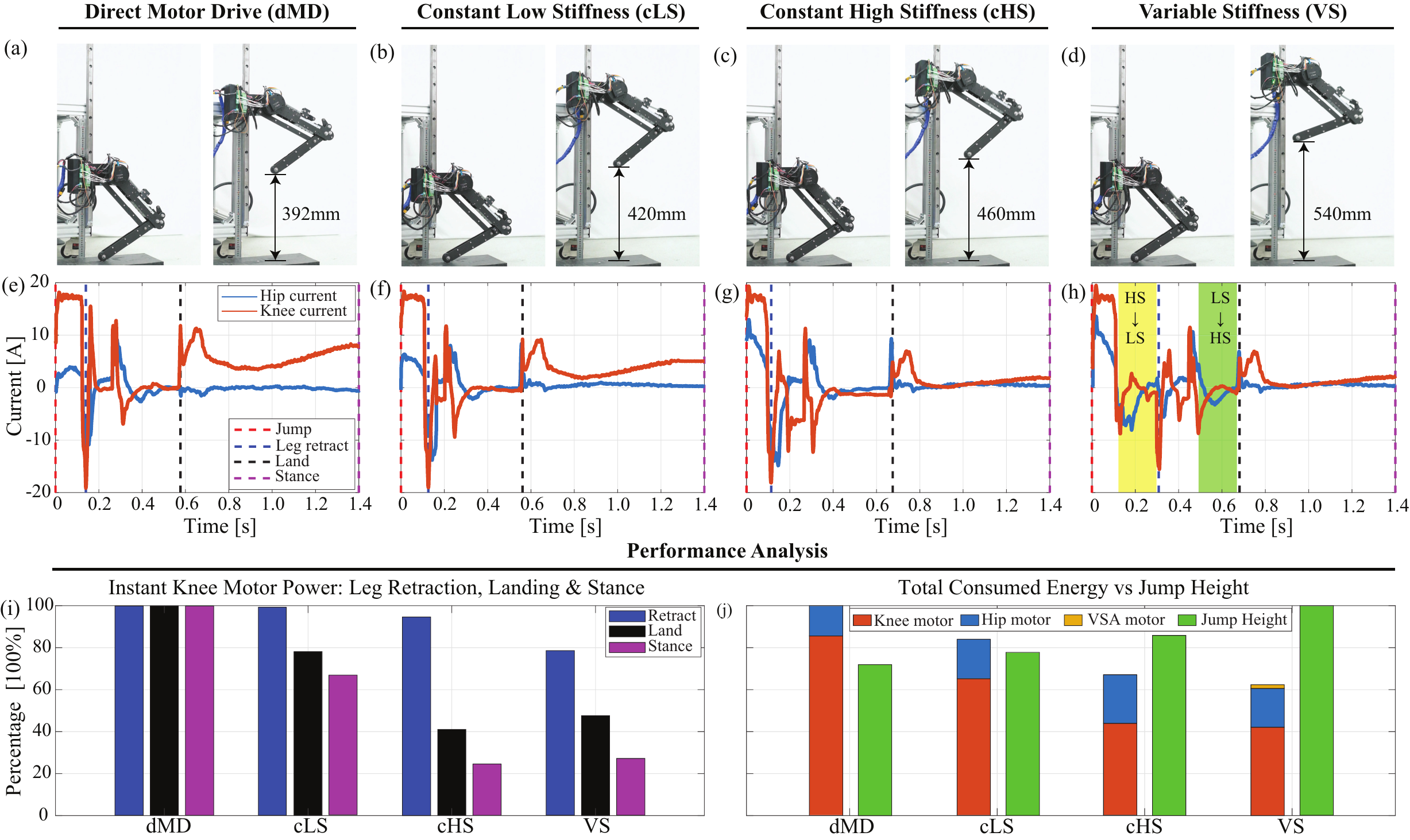}
	\caption{Experimental results. (a-d) Captures of the real-world experiment. In each sub-figure, the left capture shows the consistent initial setup and the right capture shows the moment when the foot is lifted in largest height.  (e-h) The measured currents in hip and knee motors. The jump, leg retract, land and stance moment are marked with red, blue, black and purple dashed lines. In VS mode, the stiffness change process is marked with yellow and green bars. (i-j) Performance analysis showing the instant motor power and total consumed energy compared with jump height. Control parameters are set as: $h_e^d = 500 $ mm, $h_c^d = 255$ mm, and $d_e^d=d_r^d=0$;  the feedback gains $K_p$ and $K_d$ in \eqref{vmcforce} are  set as $ \text{diag}(100,2000),\text{diag}(10,20)$ to track extension and retraction motion, $ \text{diag}(100,200),\text{diag}(10,10)$ for compliant landing. When the deflection angle of leaf-spring is $ 0^{\circ} $, the high stiffness value  is 24Nm/rad, the low stiffness value  is 11Nm/rad; the feedback gains are the same for all  modes for fair comparison.}
	\label{fig:6}
\end{figure*}

\subsection{Real-time hopping control strategy}
\label{sectionIV-B}

\textbf{Task setting:} 
Our experiments examine highly dynamic hopping tasks with active leg retraction during the entire cycle from jumping to landing,  where the scenario of such tasks is to jump over an obstacle with the leg retracted during the hopping motion. 
The legged robot's locomotion is controlled using the VMC controller defined in \eqref{eq:vmc_tq}, while the VLLSA provides supporting torque with variable stiffness using the open-loop control model in \eqref{eq1}. In the experiment, we control the robot to accelerate for the first two steps, and then in the third jump, the leg will retract to generate the best height for hopping over obstacles.

\textbf{Control strategy:} Based on the task, our hopping strategy aims to increase the distance $z_B$ between the foot and the ground, enabling the legged robot to tackle more challenging tasks in complex environments. The controller is event-triggered, starting from the initial state $\mathbf{r}_0=[d_0,h_0]^\top$ with the VLLSA in high stiffness to conserve raising power. When the jump command is given, the robot is controlled to reach the predefined extension state $\mathbf{r}_e^d=[d_e^d,h_e^d]^\top$. Once $h_e^d$ is reached, the VLLSA switches to low stiffness for easier operation, and the leg contracts to a predefined retraction state $\mathbf{r}_r^d=[d_r^d,h_r^d]^\top$, where $h_r^d<h_e^d$, to across the obstacle. Then, the tracking height extends back to $h_e^d$ and the VLLSA switches to high stiffness to provide extra support torque for landing. The hopping control strategy in VS mode is illustrated in Fig. \ref{fig:5}b, with movement of the VLLSA slider marked by red and blue arrows. 
We note that the proposed controller \eqref{controller} could be computed, at each time step, within 30us on a STM32F407IGH6 chip, CPU 168MHz with 196KB RAM,  enabling the real-time control for VLLSA-leg. 

We note that most of the prior hopping experiments with VSAs \cite{Hung2016,Guenther2019,vanderborght2011maccepa,Nicholson} focused on studying the natural dynamics with passive stiffness modulation, while our experiments adopt the active stiffness variation control to improve the hopping locomotion towards more agile dynamic performance. 


%

\section{Hopping experiment and results}
\label{section4}

\subsection{In-place hopping} \label{experi_set}

 To simplify the hopping motion and focus on investigating the  performance of the proposed VLLSA in legged robot, we first conducted an in-place hopping experiment. The hip of the legged robot is mounted on a linear guide to ensure  the hip moving vertically in a linear pattern. The hardware setting is shown in Fig. \ref{fig:6}a-d.

To highlight the advantages of VLLSA, we design the hopping experiments with four  actuation modes: 1) Direct Motor Drive (dMD) where the VLLSA is disconnected with the knee joint; 2) Constant Low Stiffness (cLS) where the slider is kept at low stiffness position; 3) Constant High Stiffness (cHS) where the slider is kept at high stiffness position; and 4) Variable Stiffness (VS) where the slider is controlled to move actively and output torque with variable stiffness. The control strategy of VLLSA in VS mode is presented in Section \ref{sectionIV-B}. 
To keep the consistency across different actuation methods, all the hopping tests are initialized with the same initial state and the same control parameters in VMC controller \eqref{eq:vmc_tq}. The experimental results are presented, analyzed in Fig. \ref{fig:6}, and recorded in the supplementary video.

\subsubsection{Hopping performance}
As the task is to imitate jumping over obstacles, the hopping performance is evaluated by the lifting height of the foot. 
The observed hopping heights are presented in Fig. \ref{fig:6}a-d for the four actuation methods. We note that i) dMD mode achieves the lowest height due to the lack of extra support from leaf spring, ii) although the cLS mode could only provide a small output torque with low stiffness, it could help jump higher than dMD, and iii) the cHS jumps higher than cLS as higher output stiffness could provide more kinetic energy accumulation. Such phenomenon coincides with the previous works that the compliant actuators in legged robot could improve the hopping performance \cite{Nan2022,Xin2015}. The VS mode achieves the highest jumping: 37.8\% higher than the dMD because the high knee stiffness could accumulate more kinetic energy, and 17.4\% higher than the cHS because the low stiffness in flight could ease the leg retraction. Consequently, it is demonstrated that the legged robot with compliant VLLSA could improve the performance in dynamic hopping tasks.


\subsubsection{Dynamic motion analysis}
The measured currents during hopping in hip and knee motors are displayed in Fig. \ref{fig:6}e-h. We define the duration between jumping (red dashed line), leg retraction (blue dashed line), and landing (black dashed line) as $t_{JL}$, $t_{LL}\in\mathbb{R}_+$. 
Comparing the dMD and cLS in Fig.\ref{fig:6}e,f, we note that the $t_{JL}$, $t_{LL}$ in dMD and cLS are almost identical, because the stiffness in cLS mode is quite low and has little influence on the hoping motion. 
Comparing the cLS and cHS in Fig. \ref{fig:6}f,g, we note that $t_{JL}^{cHS}<t_{JL}^{cLS}$ due to the larger output stiffness and higher acceleration; also, it is observed that $t_{LL}^{cHS}>t_{LL}^{cLS}$, as the knee motor need to overcome the large stiffness in cHS mode to retract the leg, so the high stiffness in knee is undesirable when retracing the leg. 

Comparing the cHS and VS in Fig. \ref{fig:6}g,h, we note that $t_{JL}^{VS}$ is approximately the sum of $t_{JL}^{cHS}$ and $t_{HS\rightarrow LS}$: the jumping process to $h_e$ is also the same between cHS and VS, as they both drive in high stiffness mode; in the air, the robot waits for the change to low stiffness to ease the leg retraction, thus extending the $t_{JL}^{VS}$. Also, we observed that $t_{LL}^{VS}$ is close to $t_{LL}^{cLS}$, as the output stiffness when retracting leg is the same as cLS mode to save energy consumption. In summary, the VS mode integrates the advantages of high stiffness when jumping and low stiffness when retracting, providing the best performance when compared to other alternative modes.


\subsubsection{Instant motor power} \label{instant_power_1}
The peak instant motor power marks the largest power supplied by the motor to the system, requiring large power output might damage the motor or deteriorating the dynamic performance of the hopping motion; thus, we would expect the peak motor power would be reduced. During the hopping, the sharp increase of motor power appears in knee motor at jumping and leg retracting moment, the analyzed data is presented in Fig. \ref{fig:6}i. 

We first note that the instant power at leg retraction, denoted by the blue bar in Fig. \ref{fig:6}i, is decided by two factors: large stiffness at knee joint or small angular velocity of the shank would consume more power to lift the leg. Consequently, we note that dMD and cLS requires the highest instant power due to the low angular velocity when lifting, cHS requires less power due to the high velocity when retracting the leg, and VS mode requires the least power because i) the lower leg has high speed as cHS, and ii) the knee stiffness has lower stiffness.

During robot landing, we note dMD has the highest instant power (denoted by the black bar in Fig. \ref{fig:6}i) due to the lack of compliant supporting, though it has the lowest jumping height; the cLS only requires 80\% instant power compared to dMD due to the introduction of VLLSA, even with low stiffness. The instant landing power in cHS and VS is almost the same, VS is slightly higher as it has larger jumping height. Similar conclusion holds in static stance, denoted by the purple bar in Fig. \ref{fig:6}i, as the static power in stance is mainly decided by the knee stiffness: larger stiffness would require less motor power to support the legged robot.

\subsubsection{Total energy consumption}
Finally we analyze the total energy consumption during hopping. It is expected that the energy consumption could be reduced as most of the mobile robotic systems are driven by batteries, saving energy would extend the duration of robot operation. It is obvious that dMD mode consumes the most energy as all the robot motion is driven by the motors; the cLS mode saves energy to dMD as the leaf spring helps support the robot during landing and  stance; the cHS mode consumes lower power than cLS as it further saves energy during jumping and  stance by providing larger output stiffness. The VS mode changes the stiffness to further save power during leg retraction when compared to cHS; finally the VS mode provides the best hopping performance while consuming the least electric power. 


\begin{figure}[t]
	\centering
	\includegraphics[width=\linewidth]{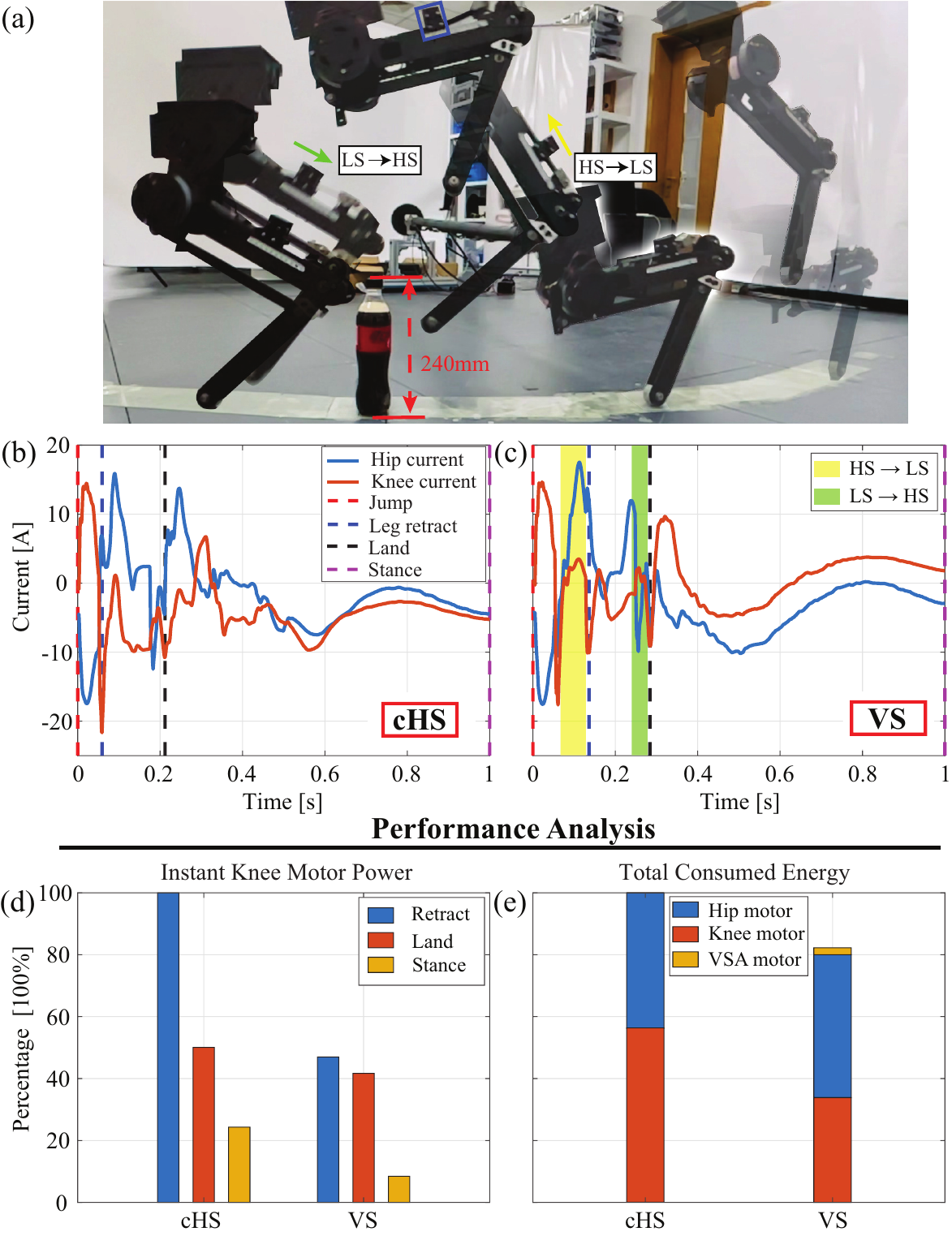}
	\caption{Experimental results. (a) Captures of the forward hopping experiment.   (b-c) Measured currents in hip and knee motors. The jump, leg retract, land and stance moment are marked with red, blue, black and purple dashed lines. In VS mode, the stiffness change process is marked with yellow and green bars. (d-e) Performance analysis showing the instant motor power and total consumed energy. Control parameters are set as: $h_e^d = 450 $ mm, $h_c^d = 250$ mm, and $d_e^d=200$ mm, $d_r^d=100$ mm;  the feedback gains $K_p$ and $K_d$ in \eqref{vmcforce} are  set as $ \text{diag}(800,800),\text{diag}(5,5)$ to track extension and retraction motion, $ \text{diag}(300,300),\text{diag}(10,10)$ for compliant landing. When the deflection angle of leaf-spring is $ 0^{\circ} $, the high stiffness value  is 65Nm/rad, the low stiffness value  is 11Nm/rad; the feedback gains are the same for all  modes for fair comparison. }
	\label{fig:7}
\end{figure}

\subsection{Forward hopping}
In this section we test the VLLLSA-leg in a more general and challenging scenario, where the robot is commanded to jump forward and over an obstacle with height of 0.24m and distance of 1.5m from the starting point. Different from the in-place hopping where we set $d_e^d=d_r^d=0$, in forward hopping we need to set $d_e^d$ and $d_r^d$ as nonzero values. We note that the forward hopping requires attention to hopping height, step length, and more significantly, maintaining proper swing posture to prevent collisions between the foot, knee, and the obstacle, making the forward hopping test a comprehensive challenge to the VLLSA-leg and the proposed controller in Section \ref{section3}. 

Similar to previous works where the forward legged hopping is tested \cite{bolignari2022diaphragm,Xin2015}, the hip of the robot is affixed to a 2m-long boom structure, as shown in Fig. \ref{fig:7}a, to avoid trunk pitch control and constrain the robot to jump along a circular trajectory over significant distances. At the other side of the boom, a counterweight is attached to balance the boom's mass, ensuring the robot is not affected by the setting. 

\subsubsection{Hopping performance}
Similar to the tests of in-place hopping, we test the actuation modes of dMD, cLS, cHS and VS. We note that, the current obstacle setting is too challenge for the dMD and cLS to even reach the location of the obstacle, due to the lack of sufficient driving torque to propel the legged robot to the intended destination and jump over the obstacle. In contrast, when employing the cHS and VS modes for hopping, the VLLSA could provide stronger supporting torque during the takeoff phase, enhancing the dynamic performance significantly. Through the accumulation of kinetic energy over two preceding hops, the legged robot successfully jumped over the obstacle in these two modes. 
Subsequently, we mainly present and analyze the performance of cHS and VS results in Fig. \ref{fig:7}b-e; for the performance of dMD and cLS modes, please refer to the supplementary video for more details.

\subsubsection{Instant motor power}

Analyzing the recorded current in hip and knee motors, we notice that the hip current is increased compared to in-place hopping, because the hip motor is required to control the robot configuration, and the hip currents between cHS and VS are around similar level; meanwhile, the current in knee motor still has larger amplitude when compared to the hip, especially in cHS mode. 
So we remain investigating  the instant power in knee motor.

During leg retraction, it is obvious that the instant knee current in VS mode is reduced compared to cHS mode, as shown in Fig. \ref{fig:7}b-c; correspondingly, the instant power is also reduced by around 50\% in VS mode (blue bar in Fig. \ref{fig:7}d). We note such decrease is more significant than the in-place hopping (around 20\%), as we set larger stiffness to support the forward hopping motion, and the knee motor would output more torque to overcome the high stiffness in knee joint during leg retraction. 

When the legged robot lands, the instant knee motor power in cHS and VS modes is similar (orange bar in Fig. \ref{fig:7}d), due to the similar jumping height and landing speed. However, at the stance to maintain a specific height, the high stiffness we set is so large that the output from VLLSA is  beyond the nominal torque of knee motor, leading the knee motor to output reverse torque and increasing the energy consumption (the knee current in Fig. \ref{fig:7}b is negative). In contrast, we set the VLLSA to actively lower the landing stiffness and balance the output torque, reducing the energy consumption by around 68\% (yellow bar in Fig. \ref{fig:7}d).



\subsubsection{Total energy consumption}
The total energy consumption during forward hopping is analyzed in Fig. \ref{fig:7}e. We note that while the energy consumed by hip motor is close in the cHS and VS modes, the main reduction in VS mode  is from the knee motor: compared to cHS mode, the VLLSA in VS mode could actively change the stiffness to save the energy consumption by 42.8\% (orange bar in Fig. \ref{fig:7}e). Moreover, the power consumed by stiffness motor is only 2.2\% of the total energy consumption. Both points indicate the energy efficiency of VLLSA in terms of modulating stiffness and optimizing agile legged locomotion. 


In summary, with the in-place and forward hopping results presented, it is clearly demonstrate that, the proposed VLLSA could actively change the output stiffness in knee joint  to integrate the advantages of low stiffness in compliant motion, like leg retraction,  and high stiffness in explosive motion, like jumping and landing. Furthermore, integration of VLLSA could reduce the  instant motor power to protect the electronic components, and save the total energy consumption to extend the operation of battery-powered mobile robots.

\section{Discussion and Conclusion}
\label{section5}
In this paper, the proposed VLLSA changes the output stiffness by changing the effective length of the leaf spring. Different from previous works, the slider to control the effective length is supported by linear guide and motor-driven ball screw, maintaining the compactness of the proposed actuator; besides, a custom made gear set and linear slider attached to the end of the leaf-spring ensures the small deflection of the spring and precise modeling of the output torque and stiffness. To fully explore the advantages of the actively controlled stiffness, a real-time controller based on VMC is proposed to integrate the control of the legged robot and the active modulation of stiffness in VLLSA. With comprehensive real-world experiments including in-place and forward hopping, it is demonstrated that the proposed VLLSA could reduce instant motor power and total energy consumption, while improve the agile locomotion performance in legged systems. 

In future work, we aim to continue optimizing the design of VLLSA to expand the stiffness variable range, possibly by stacking more pieces of springs to improve the maximum output and optimizing the shape of leaf-spring to reduce the lower bound of stiffness. Furthermore, while the proposed controller could integrate robot configuration  and output stiffness of VLLSA, we note that the stiffness is still controlled in Bang-bang type, including only two values of high and low. We aim to develop data-driven optimal control methods to enable continuous modulation of stiffness output \cite{chen2022,pratt2001virtual,Alexander2021,Chen2021}, further optimizing the hopping performance of legged robots.

\bibliographystyle{IEEEtran}
\bibliography{mylib}

\vskip -2\baselineskip plus -1fil
\begin{IEEEbiography}[{\includegraphics[width=1in,height=1.25in,clip,keepaspectratio]{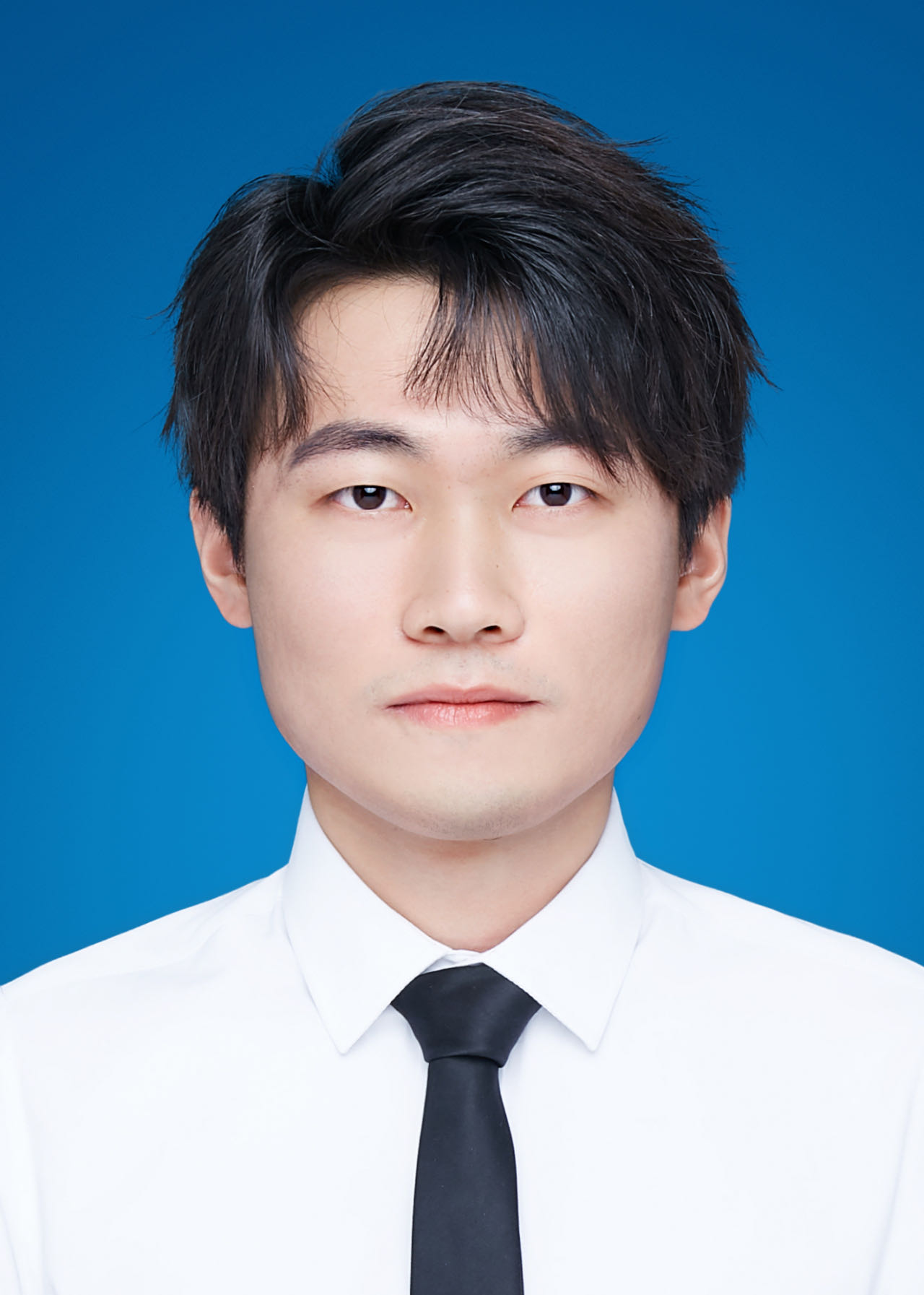}}]{\textbf{Lei Yu}} received the B.Eng. degree in Vehicle Engineering, in 2020, and the M.Res. degree in Pattern Recognition and Intelligent Systems from the University of Liverpool, Liverpool, UK, in 2022. He is currently working toward the Ph.D. degree in Electronic and Electrical Engineering with the University of Liverpool.	His research interests include variable stiffness actuator, bionic robots, and optimal control.
\end{IEEEbiography}
\vskip -2\baselineskip plus -1fil
\begin{IEEEbiography}[{\includegraphics[width=1in,height=1.25in,clip,keepaspectratio]{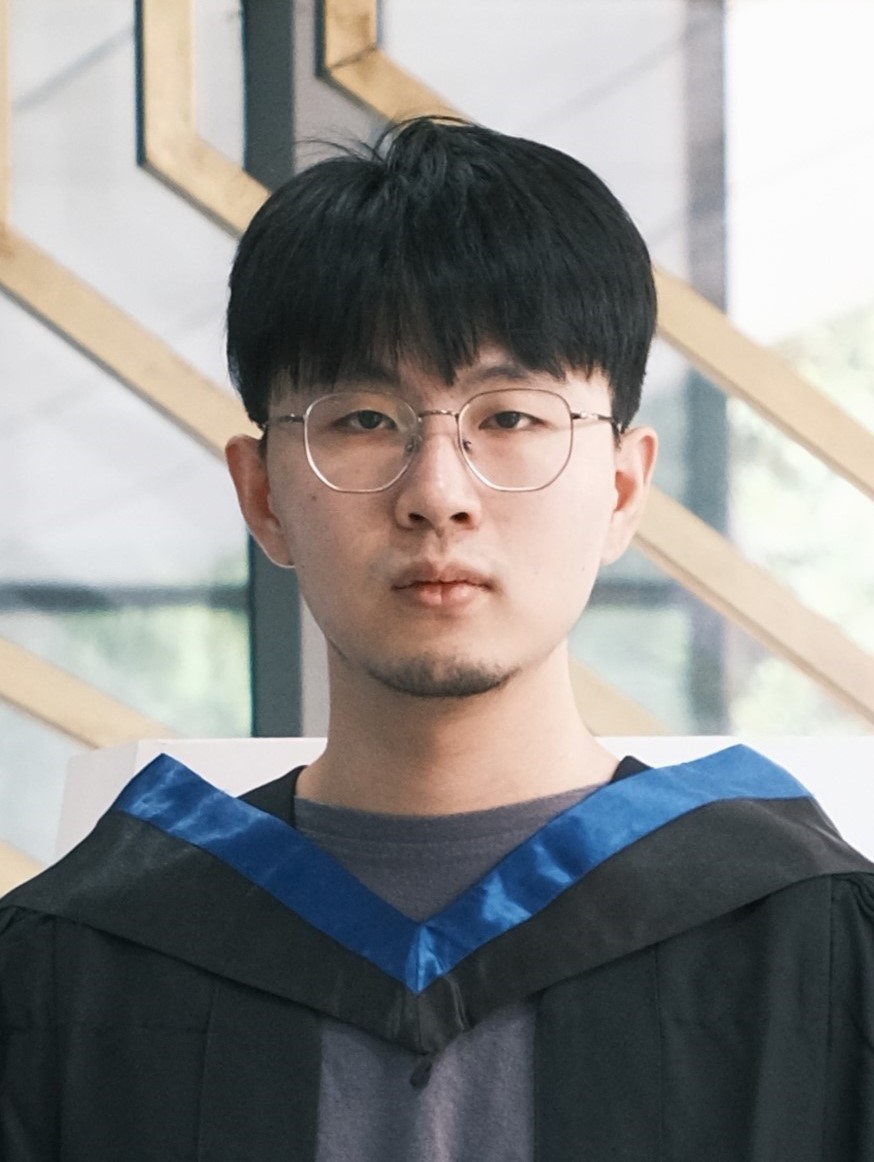}}]{\textbf{Haizhou Zhao}} received the B.Eng. degree in Mechatronics and Robotic Systems from  Xi'an Jiaotong-Liverpool University, Suzhou, China, in 2023. His research is focused on robotic control and mechanical design.
\end{IEEEbiography}
\vskip -2\baselineskip plus -1fil
\begin{IEEEbiography}[{\includegraphics[width=1in,height=1.25in,clip,keepaspectratio]{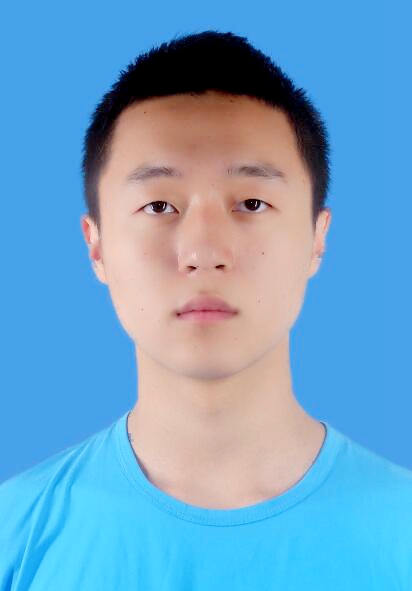}}]{\textbf{Siying Qin}} received the B.Eng. degree in Mechatronics and Robotic Systems from the Xi'an Jiaotong-Liverpool University, Suzhou,China, in 2022. He is currently working toward the Ph.D. degree in Electronic and Electrical Engineering with the University of Liverpool, Liverpool, UK. His research is focused on optimal control and legged robot.
\end{IEEEbiography}
\vskip -2\baselineskip plus -1fil
\begin{IEEEbiography}[{\includegraphics[width=1in,height=1.25in,clip,keepaspectratio]{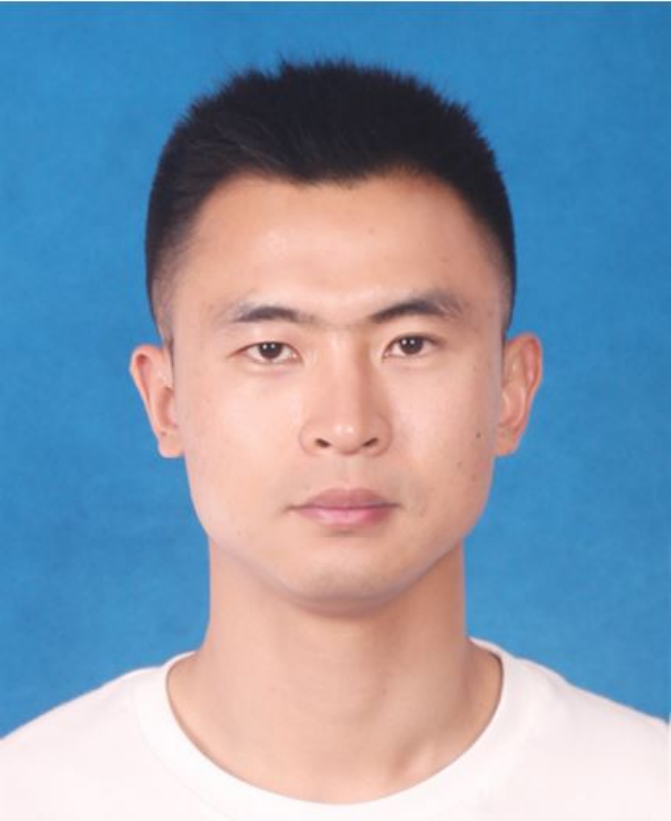}}]{Gumin Jin}
	received the B.S. degree from Harbin Institute of Technology, Harbin, China, in 2013. He is currently pursuing the Ph.D. degree with the Department of Automation, School of Electronic Information and Electric Engineering, Shanghai Jiao Tong University , China. His research interests include computer vision, 3-D sensing and metrology, and robotics.
\end{IEEEbiography}
\vskip -2\baselineskip plus -1fil
\begin{IEEEbiography}[{\includegraphics[width=1in,height=1.25in,clip,keepaspectratio]{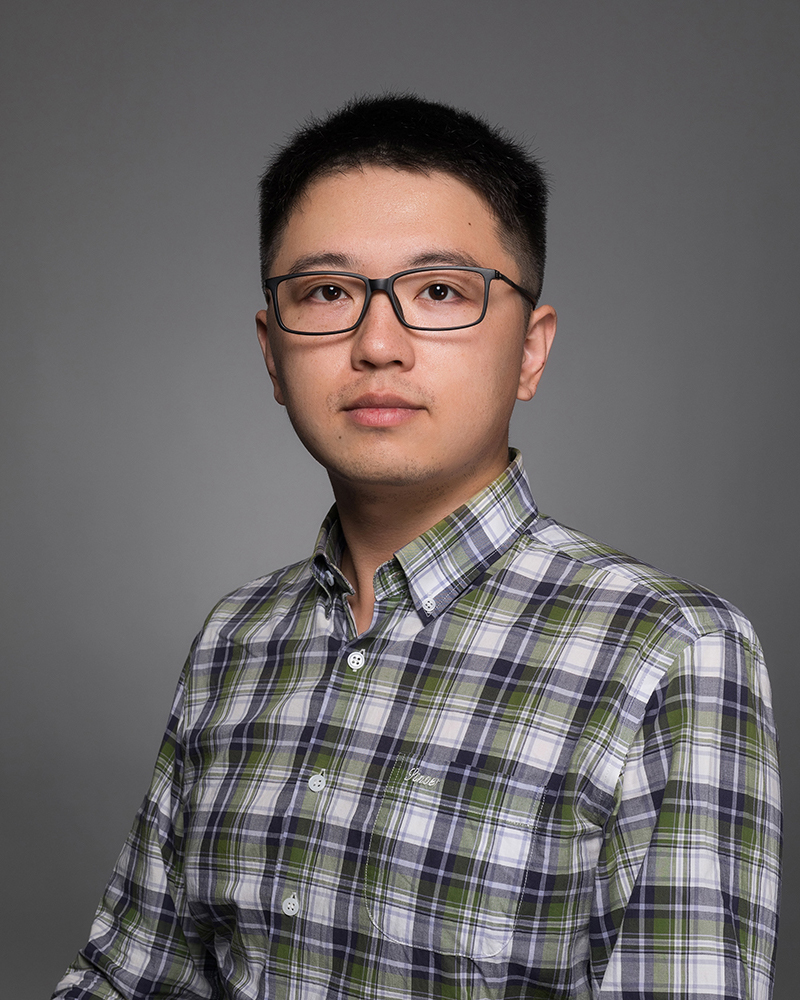}}]{\textbf{Yuqing Chen}} received his Ph.D. degree from Singapore University of Technology and Design (SUTD) in 2020, M.Eng. and B.Eng. degrees in Control Science and Engineering from Harbin Institute of Technology (HIT) in 2015 and 2013 respectively. He is currently an Assistant Professor with the Department of Mechatronics and Robotics, School of Advanced Technology, Xi'an Jiaotong-Liverpool University. His main research interests include robot control, optimal control of dynamical systems and hardware-in-the-loop optimal control theory.
\end{IEEEbiography}

\end{document}